\documentclass{bmvc2k}

\usepackage{amsmath}
\usepackage{amsfonts, amssymb}
\usepackage{relsize}
\usepackage{graphicx}
\usepackage{multirow}

\title{Foreground-aware Semantic Representations for Image Harmonization}

\addauthor{Konstantin Sofiiuk}{k.sofiiuk@samsung.com}{1}
\addauthor{Polina Popenova}{p.popenova@partner.samsung.com}{1}
\addauthor{Anton Konushin}{a.konushin@samsung.com}{1}

\addinstitution{
 Samsung AI Center\\
 Moscow, Russia
}

\runninghead{Sofiiuk \etal}{Foreground-aware Semantic Representations for harmony}

\def\eg{\emph{e.g}\bmvaOneDot}

\def\etal{\emph{et al}\bmvaOneDot}

\begin{document}

\maketitle

\begin{abstract}
Image harmonization is an important step in photo editing to achieve visual consistency in composite images by adjusting the appearances of foreground to make it compatible with background. Previous approaches to harmonize composites are based on training of encoder-decoder networks from scratch, which makes it challenging for a neural network to learn a high-level representation of objects. We propose a novel architecture to utilize the space of high-level features learned by a pre-trained classification network. We create our models as a combination of existing encoder-decoder architectures and a pre-trained foreground-aware deep high-resolution network. We extensively evaluate the proposed method on existing image harmonization benchmark and set up a new state-of-the-art in terms of~MSE and PSNR metrics. The code and trained models are available at \url{https://github.com/saic-vul/image_harmonization}.

\end{abstract}

\section{Introduction}
\label{sec:intro}
\begin{figure}[!ht]
 \centering
 \includegraphics[width=\linewidth]{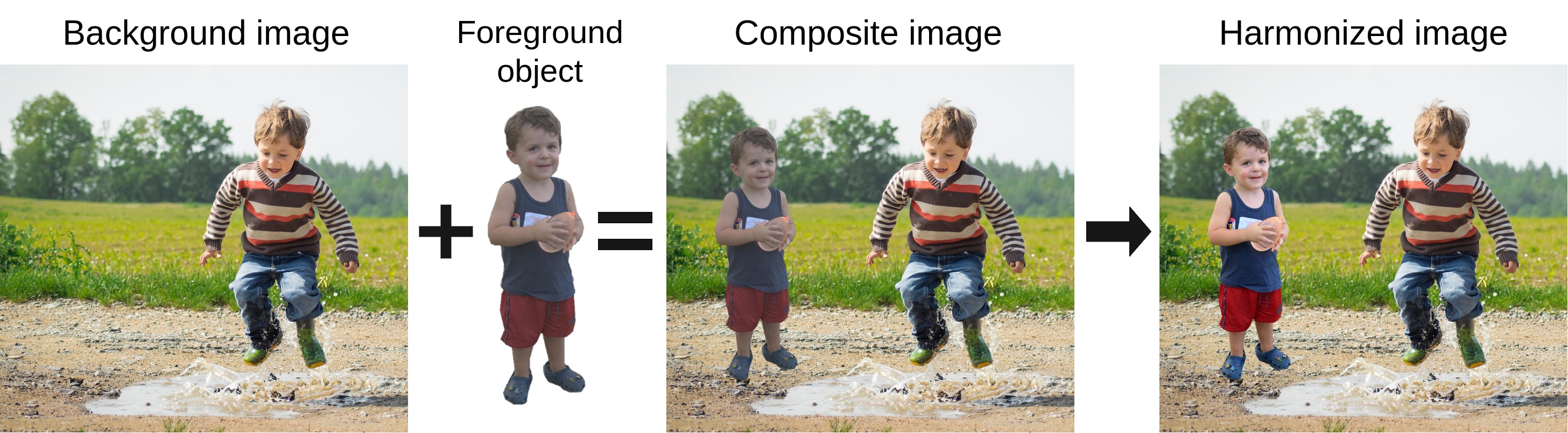}
 \caption{Illustration of the practical application of image harmonization algorithm to a~composite image. Best view in color.}   
 \label{fig:ih_explained}
\end{figure}

The main challenge of image compositing is to make the output image look realistic, given that the foreground and background appearances may differ greatly due to photo equipment specifications, brightness, contrast, etc. To address this challenge, image harmonization can be used in order to make those images visually consistent. In general, image harmonization aims to adapt the appearances of the foreground region of an image to make it compatible with the new background, as can be seen in Fig. \ref{fig:ih_explained}.

In recent years, several deep learning-based algorithms have been addressed to this problem \cite{Tsai2017DeepIH, Cun2020ImprovingTH, Cong2020DoveNetDI, Isola2017ImagetoImageTW}. Unlike traditional algorithms that use handcrafted low-level features \cite{Jia2006DraganddropP, Lalonde2007UsingCC, Sunkavalli2010MultiscaleIH, Xue2012UnderstandingAI}, deep learning algorithms can focus on the image contents.

For image harmonization, it is crucial to understand what the image foreground and background is and how they should be semantically connected. For example, if to-be-harmonized foreground object is a giraffe, it is natural to adjust the appearance and color to be blended with surrounding contents, instead of making the giraffe white or red. Therefore, Convolutional Neural Networks (CNNs) have succeeded in such tasks, showing an excellent ability to learn meaningful feature spaces, encoding diverse information ranging from low-level features to high-level semantic content~\cite{Krizhevsky2012ImageNetCW, Zhang2018InterpretableCN}.

In recent papers on image harmonization, models are trained from scratch using only annotations that do not contain any semantic information \cite{Cun2020ImprovingTH, Cong2020DoveNetDI}. The neural network is supposed to learn all dependencies between semantic patterns without supervision. However, it proves to be an extremely difficult task even for deep learning. The model is more likely to fall into a local minimum with relatively low-level patterns rather than learn high-level abstractions due to several reasons. First, learning semantic information using this approach is similar to self-supervised learning, where the task of automatic colorization is solved~\cite{Larsson2017ColorizationAA, larsson2016learning, zhang2016colorful}. However, several works focusing on self-supervised learning demonstrate that the resulting representations are still inferior to those obtained by supervised learning on the ImageNet \cite{Krizhevsky2012ImageNetCW, he2016deep}. Second, the amount of training data used for image harmonization training is by orders of magnitude smaller than the ImageNet dataset~\cite{deng2009imagenet} and other datasets used for self-supervised training. Considering all the aforementioned aspects, it seems challenging for a neural network to learn high-level semantic features from scratch only during image harmonization training. Tsai \etal \cite{Tsai2017DeepIH} also highlighted this challenge and proposed special scene parsing decoder to predict semantic segmentation, although it provided only insignificant increase in quality and required semantic segmentation annotation of all training images. Consequently, this technique was not used in the recent papers on this topic \cite{Cun2020ImprovingTH, Cong2020DoveNetDI}.

In this paper, we propose a simple approach to effective usage of high-level semantic features from models pre-trained on the ImageNet dataset for image harmonization. We find that the key factor is to transfer the image and the corresponding foreground mask to a pre-trained model. To achieve such transfer, we present a method of adapting the network to take the image and the corresponding foreground mask as the input without negatively affecting the pre-trained weights. In contrast to previous works, we make existing pre-trained models foreground-aware and combine them with encoder-decoder architectures without adding any auxiliary training tasks.

Another challenging task is to create valid training and test datasets, since producing a large dataset of real harmonized composite images requires a lot of human labor. For this reason, Tsai \etal \cite{Tsai2017DeepIH} proposed a procedure for creating synthesized datasets, although their dataset was not published. Cong \etal \cite{Cong2020DoveNetDI} reproduced this procedure, added new photos to the dataset and made it publicly available. To evaluate the proposed method, we conduct extensive experiments on this synthesized dataset and perform quantitative comparison with previous algorithms.

\section{Related work}
\label{sec:related_work}
In this section, we review image harmonization methods and some related problems as well.

\textbf{Image harmonization}. Early works on image harmonization use low-level image representations in color space to adjust foreground to background appearance. Existing methods apply alpha matting~\cite{Sun2004PoissonM, Wang2007SoftScissors}, matching color distribution~\cite{Reinhard2001ColorT, Cohen2006ColorH, Pitie2007TLinearMK} and multi-scale statistics~\cite{Sunkavalli2010MultiscaleIH}, gradient-domain methods~\cite{Perez2003PoissonIE, Jia2006DraganddropP}. Combinations of these methods are used to assess and improve realism of the images in~\cite{Lalonde2007UsingCC, Xue2012UnderstandingAI}.

Further work on image realism was provided by Zhu \etal~\cite{Zhu2015LearningDM}. They fit a CNN model to distinguish natural photographs from automatically generated composite images and adjust the color of the masked region by optimizing the predicted realism score. The first end-to-end CNN for image harmonization task was proposed by Tsai \etal~\cite{Tsai2017DeepIH}. Their Deep Image Harmonization (DIH) model exploits a well-known encoder-decoder structure with skip connections and an additional branch for semantic segmentation. The same basic structure is broadly used in the related computer vision tasks such as super resolution~\cite{Zhang2018ImageSR}, image colorization~\cite{zhang2016colorful, He2018Deep}, and image denoising~\cite{Liu2018Multilevel}. Cun \etal~\cite{Cun2020ImprovingTH} also go with an encoder-decoder U-Net-based~\cite{Ronneberger2015Unet} model and add spatial-separated attention blocks to the decoder branch.

Standard encoder-decoder architectures with a content loss can be supplemented by an adversarial loss too. Usually, these models are successfully trained with no use of the GAN structure, but in some cases adversarial learning makes an impact on image realism. The approach can be found in the papers on super resolution~\cite{Ledig2017Photo, Wang2018Fully} and image colorization~\cite{Kim2019Tag2pix, Nazeri2018Image}. Cong \etal~\cite{Cong2020DoveNetDI} construct an encoder-decoder model based on the Deep Image Harmonization architecture~\cite{Tsai2017DeepIH} with spatial-separated attention blocks~\cite{Cun2020ImprovingTH}. Besides the classic content loss between the harmonized and target image, they also add the adversarial loss from two discriminators. While the global discriminator predicts whether the given image is fake or real as usual, the domain verification discriminator checks if the foreground and background areas are consistent with each other.

\textbf{Image-to-image translation}. Image harmonization can be regarded as an image-to-image translation problem. Some GAN architectures are designed for such tasks. Isola \etal~\cite{Isola2017ImagetoImageTW} describe a pix2pix GAN, which is trained to solve image colorization and image reconstruction problems among other tasks, and can be applied to image harmonization. There are several GAN models for the related task of image blending~\cite{Wu2019GP, Zhang2020Deep}, which aims to seamlessly blend object and target images coming from different sources.

Existing general image-to-image translation frameworks are not initially designed for the image harmonization task and do not perform as well as specialized approaches. Cun \etal~\cite{Cun2020ImprovingTH} present the results for both dedicated model, based on U-Net, and pix2pix model \cite{Isola2017ImagetoImageTW}, and the latter fails to get competitive metric values.

\textbf{Single image GANs}. In order to generate realistic images, GANs require a lot of training samples. There are recent works on adversarial training with just a single image~\cite{Shaham2019Singan, Hinz2020Improved}. The SinGAN and ConSinGAN models do not require a large training set and learn to synthesize images similar to a single sample. It could also be used for unsupervised image harmonization. However, these models show good performance only in cases when the background image contains many texture and stylistic details to be learned (\eg a painting), and the foreground is photorealistic, leading to artistic image harmonization. When it comes to the composite images based on the real-world photos, the model may achieve rather poor results. Also, SinGAN models require training from scratch for every image, so harmonization of one composite may take a lot of computational time.


\section{Proposed method}
\label{sec:method}

In this section, we first revisit the encoder-decoder architectures for image harmonization in Sec. \ref{subsec:method_encoder_decoder}. Then, we introduce foreground mask fusion module to make pre-trained image classification models foreground-aware in Sec. \ref{subsec:method_foreground_aware} and demonstrate the detailed resulting network architecture in Sec. \ref{subsec:method_final_design}.  Finally, we present Foreground-Normalized MSE objective function in Sec. \ref{subsec:method_fn_mse}.

Below we use the following unified notation. We denote an input image by $I \in \mathbb{R}^{H \times W \times 3}$, a provided binary mask of the composite foreground region by $M \in \mathbb{R}^{H \times W \times 1}$, a binary mask of the background region by $\overline{M} = 1 - M$ and concatenation of $I$ and $M$ by $\hat{I} \in \mathbb{R}^{H \times W \times 4}$  where $H$ and $W$ are height and width of the image.

\subsection{Encoder-decoder networks}
\label{subsec:method_encoder_decoder}

\begin{figure}
 \centering
 \includegraphics[width=0.95\linewidth]{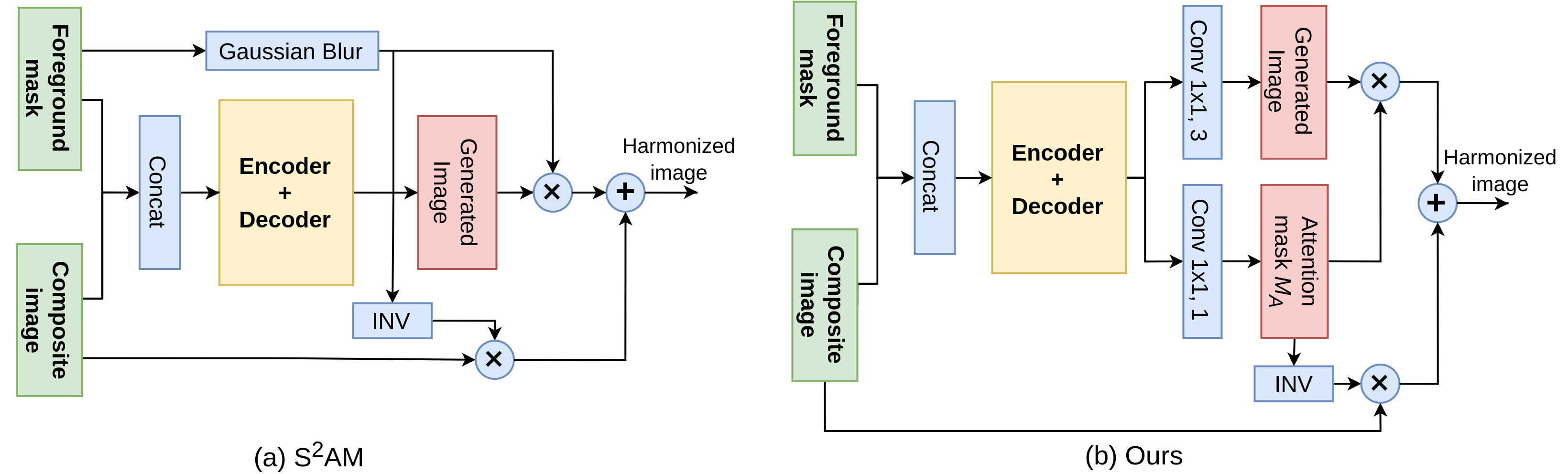}
 \caption{(a) The architecture of S$^2$AM  \cite{Cun2020ImprovingTH} approach to inference the resulting image. \newline (b) The architecture of our proposed approach. We find that the network can easily predict the mask for blending, since the foreground mask is fed as input to the encoder-decoder. No need for heuristics, \eg blurring the mask.}   
 \label{fig:fusion_comparison}
\end{figure}

We consider image harmonization as an image-to-image problem, with a key feature of mapping a high resolution input to a high resolution output. Many previous works have used an encoder-decoder network \cite{Tsai2017DeepIH, Cun2020ImprovingTH, Cong2020DoveNetDI}, where the input is passed through a series of layers that progressively downsample until a bottleneck layer, where the process is reversed. Such a network requires that all information flow passes through all the layers, including the bottleneck. Skip connections between an encoder and a decoder are essential for image harmonization, as they help to avoid image blurring and texture details missing.

In this section, we present our modification of encoder-decoder network with skip connections that was introduced in DIH \cite{Tsai2017DeepIH}. We change the original DIH architecture by removing fully connected bottleneck layer to make the model fully convolutional.

In the original DIH, the network reconstructs both the background and foreground regions of an image, which is not optimal for solving the task of image harmonization, as the background should remain unchanged. Therefore, we propose a simple scheme inspired by S$^2$AM  \cite{Cun2020ImprovingTH}, where the network predicts the foreground region and the attention mask $M_{A}$. The final image is obtained by blending the original image and the decoder output using the attention mask, as shown in Fig.~\ref{fig:fusion_comparison}.

Let us denote the encoder-decoder network as $D_F(\hat{I}): \mathbb{R}^{H \times W \times 4} \to \mathbb{R}^{H \times W \times C}$, which returns features $x=D_F(\hat{I})$ with $C$ channels, and denote two $1 \times 1$ convolution layers as $D_{RGB}: \mathbb{R}^{H \times W \times C} \to \mathbb{R}^{H \times W \times 3}$, $D_{M}: \mathbb{R}^{H \times W \times C} \to \mathbb{R}^{H \times W \times 1}$. The output image can be formalized as:

\begin{equation}
    I^{pred} = I \times [1 - D_{M}(x)] + D_{RGB}(x) \times D_{M}(x).
\end{equation}

We refer our modification of the DIH architecture with the above-mentioned enhancements as improved DIH (iDIH).

\subsection{Foreground-aware pre-trained networks}
\label{subsec:method_foreground_aware}

In many computer vision domains such as semantic segmentation, object detection, pose estimation, etc., there are successful practices of using neural networks pre-trained on the ImageNet dataset as backbones. However, for image harmonization and other image-to-image translation problems, there is no general practice of using such networks.

Similarly to image-to-image translation, semantic segmentation models should have same input and output resolution.
Thus, semantic segmentation models seem promising for image harmonization, because they typically can produce detailed high-resolution output with a large receptive field \cite{luo2016understanding}. We choose current state-of-the-art semantic segmentation architectures such as HRNet+OCR~\cite{WangSCJDZLMTWLX19, yuan2019object} and DeepLabV3+~\cite{chen2018encoder} as the base for our experiments.  

\textbf{Foreground masks for pre-trained models}. Models trained on the ImageNet take an RGB image as input. Without awareness of the foreground mask, the network will not be able to accurately compute specific features for the foreground and the background and compare them with each other, which can negatively affect the quality of prediction. Similar issues arise in interactive segmentation and RGB-D segmentation tasks \cite{xu2016deep, benenson2019large, hazirbas2016fusenet}. The most common solution is to augment the weights of the first convolution layer of a pre-trained model to accept N-channels input instead of only an RGB image. We discover that this solution can be modified by adding an extra convolutional layer that takes the foreground mask as an input and produces 64 output channels (to match conv1 in pre-trained models). Then, its outputs are summarized with the conv1 layer outputs of the pre-trained model. A core feature of this approach is that it allows setting a different learning rate for the weights that process the foreground mask.

\subsection{Final architecture design}
\label{subsec:method_final_design}

We extract only high-level features from the pre-trained models: outputs after ASPP block in DeepLabV3+~\cite{chen2018encoder} and outputs after OCR module in HRNet~\cite{WangSCJDZLMTWLX19, yuan2019object}. There are several ways to pass the extracted features to the main encoder-decoder block:
\begin{itemize}
    \item pass them to one fixed position of the encoder (or decoder or both simultaneously);
    \item build a feature pyramid \cite{lin2017feature, WangSCJDZLMTWLX19} on them and pass the resulting features maps to the encoder layers with appropriate resolutions (or decoder or both simultaneously).
\end{itemize}

\begin{figure}[t]
 \centering
 \includegraphics[width=0.97\linewidth]{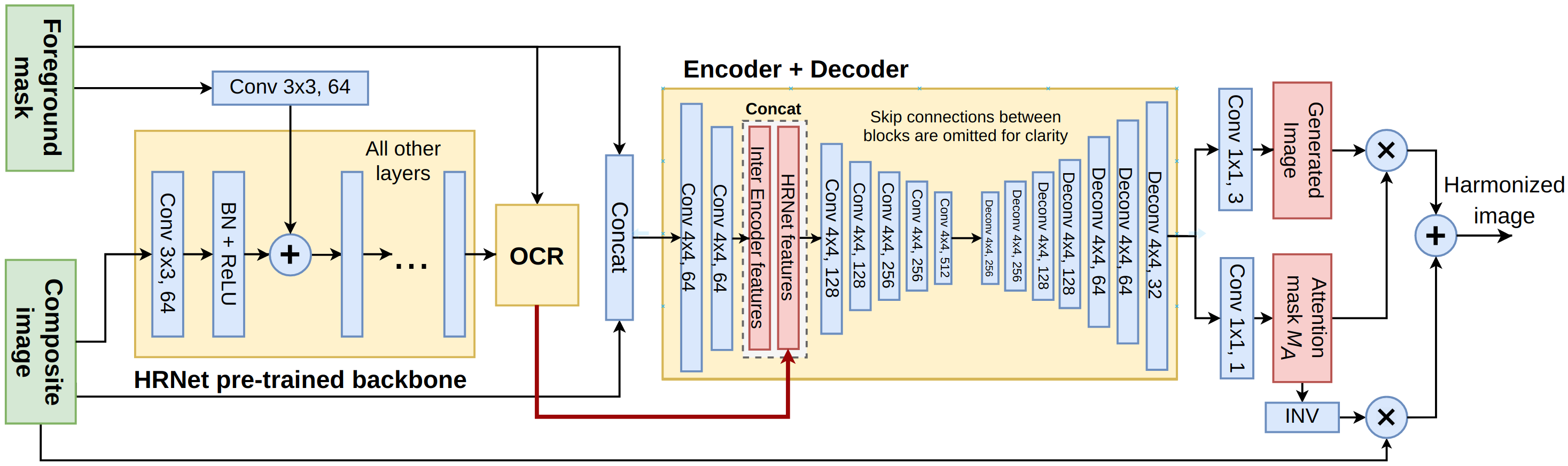}
 \caption{Our final architecture based on HRNet+OCR \cite{yuan2019object} (iDIH-HRNet).}   
 \label{fig:final_architecture}
\end{figure}

Our experiments have shown that the best results are obtained by passing the features only to the encoder. Surprisingly, when passing the features only to the decoder the worst results are obtained. It can be explained by the fact that high-level reasoning in the encoder-decoder happens at the bottleneck point, then the decoder relies on this information and progressively reconstructs the output image. Simultaneous passing of the features to both the encoder and the decoder is not reasonable due to usage of skip connections between them.

We implement a variant of feature pyramid as in HRNetV2p \cite{WangSCJDZLMTWLX19} and find that there is no need to use it to pass the features to the encoder. Without any loss of accuracy, it is sufficient to pass the features to a single position in the encoder, since the encoder is trained from scratch and can model a feature pyramid by its own structure. We aggregate the extracted features with the intermediate encoder feature map with concatenation.

Figure \ref{fig:final_architecture} shows a detailed diagram of our architecture based on HRNet+OCR \cite{yuan2019object}. We further refer to it as iDIH-HRNet, additionally specifying the backbone width. 

\subsection{Foreground-normalized MSE}
\label{subsec:method_fn_mse}

As we mentioned in section \ref{subsec:method_encoder_decoder}, the characteristic of image harmonization task is that the background region of the output image should remain unchanged relatively to the input composite. When the network takes the foreground mask as an input, it learns to copy the background region easily. Hence, the pixel-wise errors in this region will become close to zero during training. This means that training samples with foreground objects of different sizes will be trained with different loss magnitudes, which leads to poor training on images with small objects. To address this issue, we propose a modification of the MSE objective function, which is normalized by the foreground object area:

\begin{equation}
     \mathcal{L}_{rec}(\hat{I}) = \frac{1}{\max\Big\{ A_{min}, \mathlarger{\sum}\limits_{h, w}{M_{h, w}} \Big\}} \mathlarger{\sum}_{h, w} \left \| I^{pred}_{h, w} -  I_{h, w}\right \|_2^2,
\end{equation}

where $A_{min}$ is a hyperparameter that prevents instability of the loss function on images with very small objects. In all our experiments, we set $A_{min} = 100$.

\section{Experiments}
\label{sec:experiments}

\begin{figure}[t]
 \centering
 \includegraphics[width=\linewidth]{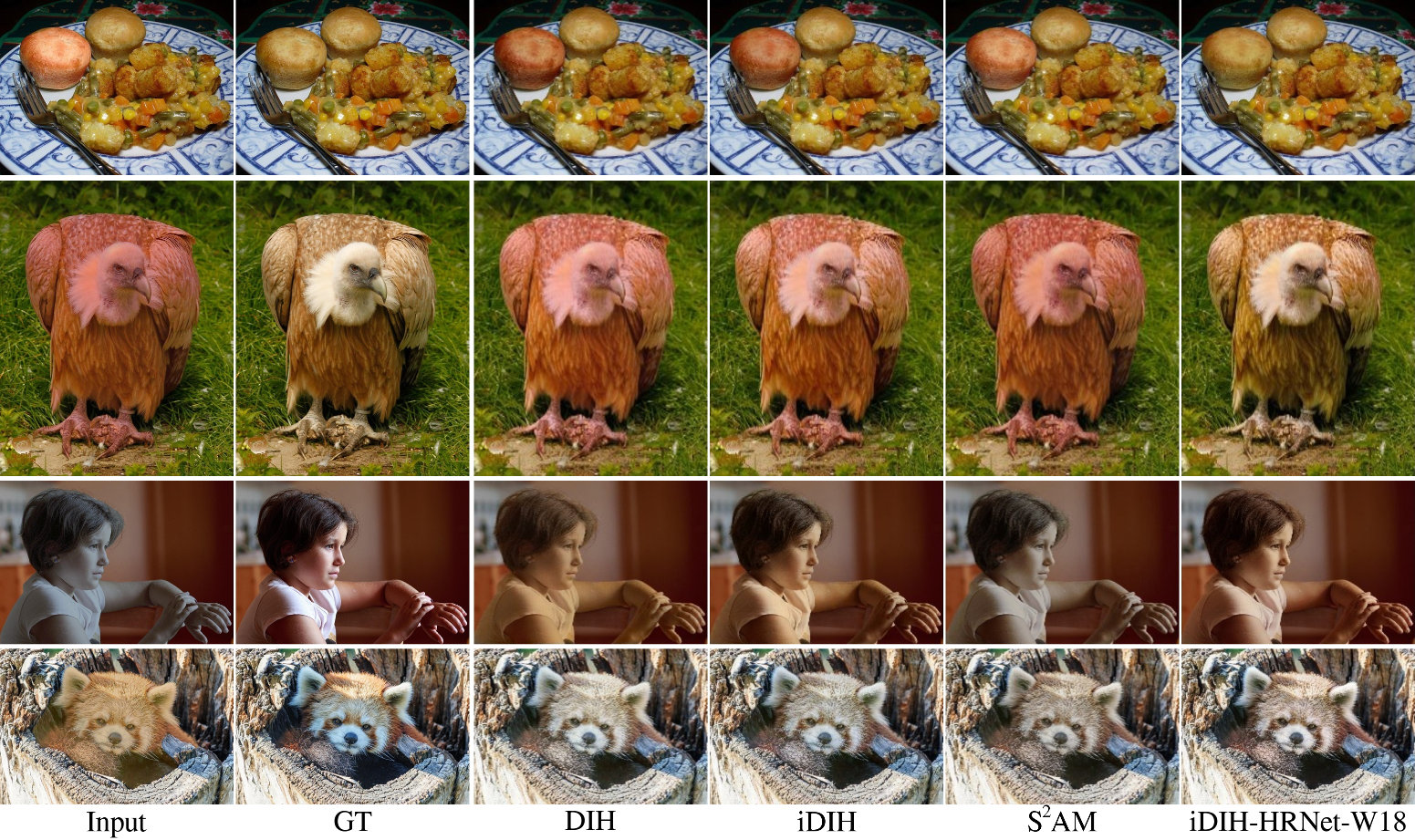}
 \caption{Qualitative comparison of different models using samples from the iHarmony4 test set. All presented models are trained with horizontal flip, RRC augmentations, and FN-MSE objective function. More
results in supplementary material.}   
 \label{fig:final_vis_comparison}
\end{figure}

\subsection{Datasets}
We use iHarmony4 dataset contributed by~\cite{Cong2020DoveNetDI}, which consists of four subdatasets:

\textbf{HCOCO} is synthesized from the joined training and test sets of COCO~\cite{Lin2014MicrosoftCoco}. The composite image in HCOCO consists of the background and foreground areas both drawn from the real image, although the foreground appearance is modified. The foreground color information is transferred from another image foreground belonging to the same category. HCOCO subdataset contains 38545 training and 4283 test pairs of composite and real images. 

\textbf{HFlickr} is based on 4833 images crawled from Flickr with one or two manually segmented foreground objects. The process of the composite image construction is the same as for HCOCO. HFlickr subdataset contains 7449 training and 828 test pairs of composite and real images.  

\textbf{HAdobe5k} is based on MIT-Adobe FiveK dataset~\cite{Bychkovsky2011Learning}. 4329 images with one manually segmented foreground object are used to build HAdobe5k subdataset. Background of the composite image is retrieved from the raw photo and the foreground is retrieved from one of the 5 edited images. HAdobe5k subdataset contains 19437 training and 2160 test pairs of composite and real images. 

\textbf{Hday2night} is based on Day2night dataset~\cite{Zhou2016Evaluating}. 106 target images from 80 scenes with one manually segmented foreground object are selected to synthesize Hday2night subdataset. The process of the composite image construction is the same as for HAdobe5k. The background and foreground images are taken from different pictures of one scene. Hday2night subdataset contains 311 training and 133 test pairs of composite and real images.

\subsection{Training details}

We use the PyTorch framework to implement our models. The models are trained for 180 epochs with Adam optimizer~\cite{Kingma2014Adam} with $\beta_1 = 0.9$, $\beta_2=0.999$ and $\varepsilon = 10^{-8}$. Learning rate is initialized with $0.001$ and reduced by a factor of $10$ at epochs 160 and 175.
The semantic segmentation backbone weights are initialized with weights from the models pre-trained on the ImageNet and updated at every step with learning rate multiplied by $0.1$. All models are trained on the iHarmony4 training set, which encapsulates training sets from all four subdatasets.

We resize input images as $256 \times 256$ during both training and testing. The input images are scaled to $[0; 1]$ and normalized with RGB mean $(0.485, 0.456, 0.406)$ and standard deviation $(0.229, 0.224, 0.225)$. Training samples are augmented with horizontal flip and random size crop with the size of the cropped region not smaller than the halved input size. The cropped image is resized to $256 \times 256$ then.

\subsection{Comparison with existing methods}

The results of traditional methods are demonstrated in previous works~\cite{Tsai2017DeepIH, Cong2020DoveNetDI}, showing that deep learning approaches perform generally better, so we compare our method with them only. We implement two baseline methods, S$^2$AM~\cite{Cun2020ImprovingTH} and DIH~\cite{Tsai2017DeepIH} without segmentation branch.

We provide MSE and PSNR metrics on the test sets for each subdataset separately and for combination of datasets in Table~\ref{tab:main_results}. Following~\cite{Cong2020DoveNetDI}, we study the impact of the foreground ratio on the model performance. In order to do that, we introduce foreground MSE (fMSE) metric which computes MSE for the foreground area only. The models metrics on the whole test set across three ranges of foreground ratios are provided in Table \ref{tab:fg_ratio}. The results show that our final model outperforms the baselines not only on the whole test set, but also on each foreground ratio.

\begin{table*}[t]
\scriptsize
\begin{center}
\begin{tabular}{l|c|c|c|c|c|c|c|c|c|c}
\hline

\multirow{2}{*}{Method} & \multicolumn{2}{c|}{HCOCO} & \multicolumn{2}{c|}{HAdobe5k} & \multicolumn{2}{c|}{HFlickr} & \multicolumn{2}{c|}{Hday2night} & \multicolumn{2}{c}{All} \\
\cline{2-11}
{}                     & MSE   & PSNR  & MSE   & PSNR  & MSE   & PSNR  & MSE   & PSNR  & MSE   & PSNR \\
\hline
\hline

DIH \cite{Tsai2017DeepIH, Cong2020DoveNetDI}
                       & 51.85 & 34.69 & 92.65 & 32.28 & 163.38 & 29.55 & 82.34 & 34.62 & 76.77 & 33.41 \\

S$^2$AM \cite{Cun2020ImprovingTH, Cong2020DoveNetDI}
                       & 41.07 & 35.47 & 63.40 & 33.77 & 143.45 & 30.03 & 76.61 & 34.50 & 59.67 & 34.35 \\
                       
DoveNet \cite{Cong2020DoveNetDI}
                       & 36.72 & 35.83 & 52.32 & 34.34 & 133.14 & 30.21 & 54.05 & 35.18 & 52.36 & 34.75 \\
\hline
DIH+augs+FN-MSE        & 22.46 & 37.85 & 36.29 & 35.39 & 93.93 & 31.97 & \textbf{40.06} & 36.89 & 34.80 & 36.46 \\
iDIH+augs+FN-MSE       & 19.30 & 38.43 & 31.33 & 36.01 & 86.20 & 32.55 & 47.18 & 37.12 & 30.79 & 37.05 \\
iDIH-HRNet-W18         & \textbf{13.93} & \textbf{39.63} & \textbf{21.80} & \textbf{37.19} & \textbf{59.42} & \textbf{33.88} & 60.18 & \textbf{37.71} & \textbf{22.15} & \textbf{38.24} \\

\hline
\end{tabular}
\end{center}
\caption{Performance comparison between methods on the iHarmony4 test sets. The best results are in bold.}
\label{tab:main_results}
\end{table*}

\begin{table*}[t]
\footnotesize
\begin{center}
\begin{tabular}{l|c|c|c|c|c|c|c|c}
\hline

Foreground ratios & \multicolumn{2}{c|}{$0\%\sim 5\%$} & \multicolumn{2}{c|}{$5\%\sim 15\%$} & \multicolumn{2}{c|}{$15\%\sim 100\%$} & \multicolumn{2}{c}{$0\%\sim 100\%$} \\
\hline
{}                     & MSE$\downarrow$   & fMSE$\downarrow$  & MSE$\downarrow$   & fMSE$\downarrow$  & MSE$\downarrow$  & fMSE$\downarrow$  & MSE$\downarrow$   & fMSE$\downarrow$  \\
\hline
\hline

DIH \cite{Tsai2017DeepIH, Cong2020DoveNetDI} 
                  & 18.92 & 799.17 & 64.23 & 725.86 & 228.86 & 768.89 & 76.77 & 773.18 \\
S$^2$AM \cite{Cun2020ImprovingTH, Cong2020DoveNetDI}
                  & 15.09 & 623.11 & 48.33 & 540.54 & 177.62 & 592.83 & 59.67 & 594.67 \\
DoveNet \cite{Cong2020DoveNetDI}
                  & 14.03 & 591.88 & 44.90 & 504.42 & 152.07 & 505.82 & 52.36 & 549.96 \\
\hline
DIH+augs+FN-MSE   & 9.45  & 399.57 & 29.51 & 329.30 & 101.34 & 331.53 & 34.80 & 366.17 \\
iDIH+augs+FN-MSE  & 8.48  & 371.47 & 25.85 & 294.64 & 89.68  & 296.80 & 30.79 & 334.89 \\
iDIH-HRNet-W18    & \textbf{6.79}  & \textbf{296.18} & \textbf{19.43} & \textbf{222.49} & \textbf{61.85}  & \textbf{202.80} & \textbf{22.15} & \textbf{256.34} \\

\hline
\end{tabular}
\end{center}
\caption{MSE and foreground MSE (fMSE) of different methods in each foreground ratio range based on the whole iHarmony4 test set. The best results are in bold.}
\label{tab:fg_ratio}
\end{table*}

\subsection{Ablation studies}

Our modifications of the baseline methods generally improve harmonization. The detailed ablation studies are shown in Table \ref{tab:ablations}. First, we upgrade training process for the DIH model by adding horizontal flip and random size crop augmentations to increase the diversity of the training data. The additional blending layer preserving the background details is then added to DIH resulting in the iDIH model. With these adjustments, the DIH architecture performs significantly better than the original one, so they are preserved in all further experiments. We compare MSE and FN-MSE objective functions on DIH and iDIH models. The use of~FN-MSE shows a consistent improvement in metrics.

We conduct experiments on three HRNet settings: HRNetV2-W18 and HRNetV2-W32 with 4 high-resolution blocks, HRNetV2-W18s with 2 high-resolution blocks. We observe that increase of HRNet complexity slightly increases quantitative results. In addition, we conduct an experiment with ResNet-34 \& DeepLabV3+~\cite{chen2018encoder}, but do not notice any improvements regarding HRNet.

We observe that features from the foreground-aware pre-trained networks increase quality for any model they are incorporated into, while the models with no pre-trained semantic information show rather poorer performance.
When training an encoder-decoder model from scratch, we need it to learn a lot of semantic features. Therefore, it is beneficial to increase the capacity of the model sacrificing its simplicity. However, simpler architecture can be used with the pre-trained networks already containing necessary information.
The iDIH structure is much lighter than the S$^2$AM, which training takes 104 hours on single GTX 2080 Ti, while iDIH training requires just 13 hours.
Considering that S$^2$AM model requires much longer training time and tends to get metric values inferior to the iDIH model, we proceed with the latter.

\begin{table*}[t]
\footnotesize
\begin{center}
\begin{tabular}{l|c|c|c|c|c|c}
\hline

Model & HFlip & RRC  & FN-MSE &   MSE$\downarrow$   & fMSE$\downarrow$  & PSNR$\uparrow$ \\
\hline
\hline
DIH                   &  --   &  --  &  --    &  65.65  & 632.86 & 34.13 \\
DIH                   &  +    &  --  &  --    &  55.44  & 555.92 & 34.73 \\
DIH                   &  +    &  +   &  --    &  36.97  & 398.26 & 35.92 \\
DIH                   &  +    &  +   &  +     &  34.80  & 366.17 & 36.46 \\
\hline
iDIH                  &  +    &  +   &  --    &  33.96  & 378.25 & 36.53 \\
iDIH                  &  +    &  +   &  +     &  30.79  & 334.89 & 37.05 \\
S$^2$AM               &  +    &  +   &  --    &  30.01  & 321.97 & 36.71 \\
\hline
iDIH-HRNet-W18s       &  +    &  +   &  +     &  22.34  & 256.78 & 38.24 \\
\begin{tabular}{@{}l@{}}iDIH-HRNet-W18s \\[-1.5mm] \scriptsize{non foreground-aware HRNet}\end{tabular}
                      &  +    &  +   &  +     &  27.00  & 300.14 & 37.58 \\
\begin{tabular}{@{}l@{}}iDIH-HRNet-W18s \\[-1.5mm] \scriptsize{pass features using feature pyramid}\end{tabular}
                      &  +    &  +   &  +     &  23.07  & 261.77 & 38.13 \\
iDIH-HRNet-W18        &  +    &  +   &  +     &  \textbf{22.15}  & \textbf{256.34} & \textbf{38.24} \\
iDIH-HRNet-W32        &  +    &  +   &  +     &  23.67  & 268.26 & 38.18 \\
iDIH-DeepLabV3+ (R-34)&  +    &  +   &  +     &  26.87  & 302.60 & 37.54 \\
S$^2$AM-HRNet-W18s    &  +    &  +   &  +     &  23.27  & 253.10 & 38.18 \\
S$^2$AM-HRNet-W18     &  +    &  +   &  +     &  24.67  & 260.70 & 37.95 \\

\hline
\end{tabular}
\end{center}
\caption{Ablation studies of different augmentation strategies, the proposed modifications of our method, and the proposed FN-MSE objective function. "HFlip" stands for horizontal flip and "RRC" stands for RandomResizedCrop augmentation.}
\label{tab:ablations}
\end{table*}
\section{Conclusion}
\label{sec:conclusion}

We propose a novel approach to incorporating high-level semantic features from models pre-trained on the ImageNet dataset into the encoder-decoder architectures for image harmonization. We also present a new FN-MSE objective function proven to be effective for this task. 
Furthermore, we observe that the use of simple training augmentations significantly improves the performance of the baselines. Experimental results show that our method is considerably superior to existing approaches in terms of MSE and PSNR metrics.

\bibliography{egbib}
\end{document}